\documentclass[letterpaper, 10 pt, conference]{ieeeconf}  

\IEEEoverridecommandlockouts                              

\usepackage{xcolor} 
\usepackage{graphicx}
\usepackage{lipsum}
\usepackage{multicol}
\usepackage{dialogue}
\usepackage{float}

\title{\LARGE \bf
ARDIE: AR, Dialogue, and Eye Gaze Policies for Human-Robot Collaboration
}
\author{Chelsea Zou, Kishan Chandan, Yan Ding, Shiqi Zhang}
\begin{document}
\maketitle

\begin{abstract}
Human-robot collaboration (HRC) has become increasingly relevant in industrial, household, and commercial settings. However, the effectiveness of such collaborations is highly dependent on the human and robots' situational awareness of the environment. Improving this awareness includes not only aligning perceptions in a shared workspace, but also bidirectionally communicating intent and visualizing different states of the environment to enhance scene understanding. In this paper, we propose ARDIE (Augmented Reality with Dialogue and Eye Gaze), a novel intelligent agent that leverages multi-modal feedback cues to enhance HRC. Our system utilizes a decision theoretic framework to formulate a joint policy that incorporates interactive augmented reality (AR), natural language, and eye gaze to portray current and future states of the environment. Through object-specific AR renders, the human can visualize future object interactions to make adjustments as needed, ultimately providing an interactive and efficient collaboration between humans and robots. 
\end{abstract}

\section{INTRODUCTION}
The ability for robots to collaborate effectively with humans is highly dependent on a mutual scene perception and shared situational understanding \cite{dos2020situational}.
This requires not only the human and robot to understand and interpret each other's intentions, but also convey important contextual information to enhance scene understanding. This is typically challenging because humans and robots prefer to communicate information differently \cite{chandan2021arroch}, \cite{bonarini2020communication}.
While humans communicate knowledge using a variety of modalities such as speech, gaze, and gestures \cite{green2007augmented}, robots tend to rely on digital information like text-based commands or visual images \cite{chandanlearning}. However, these modes of communication by themselves may not always be sufficient or clear enough to convey intentions and goals effectively. In order to understand the other party, a shared situational awareness and perceptual understanding of the scene must hold between the human and robot.

Augmented reality (AR) technologies have been developed for a wide range of applications such as immersive gaming, architectural design, educational training, and more~\cite{chen2019overview}. Recently, AR has gained significant traction in human-robot collaboration (HRC) domains for its capability to provide a shared perceptual medium between humans and robots~\cite{azuma2001recent,muhammad2019creating}.
By overlaying digital information onto physical objects, the human can interact with a digital augmented scene representation. Several studies have demonstrated the benefits of AR through its capacity to model objects as interactive scene elements~\cite{weisz2017assistive,zhang2019vision,ghiringhelli2014interactive}, and more recently, AR has also been utilized to produce visualizations of internal states of robots, enabling humans to leverage the information to adapt robot behaviors accordingly~\cite{chandan2021arroch}. 
\begin{figure}[t]
\includegraphics[width=\columnwidth]{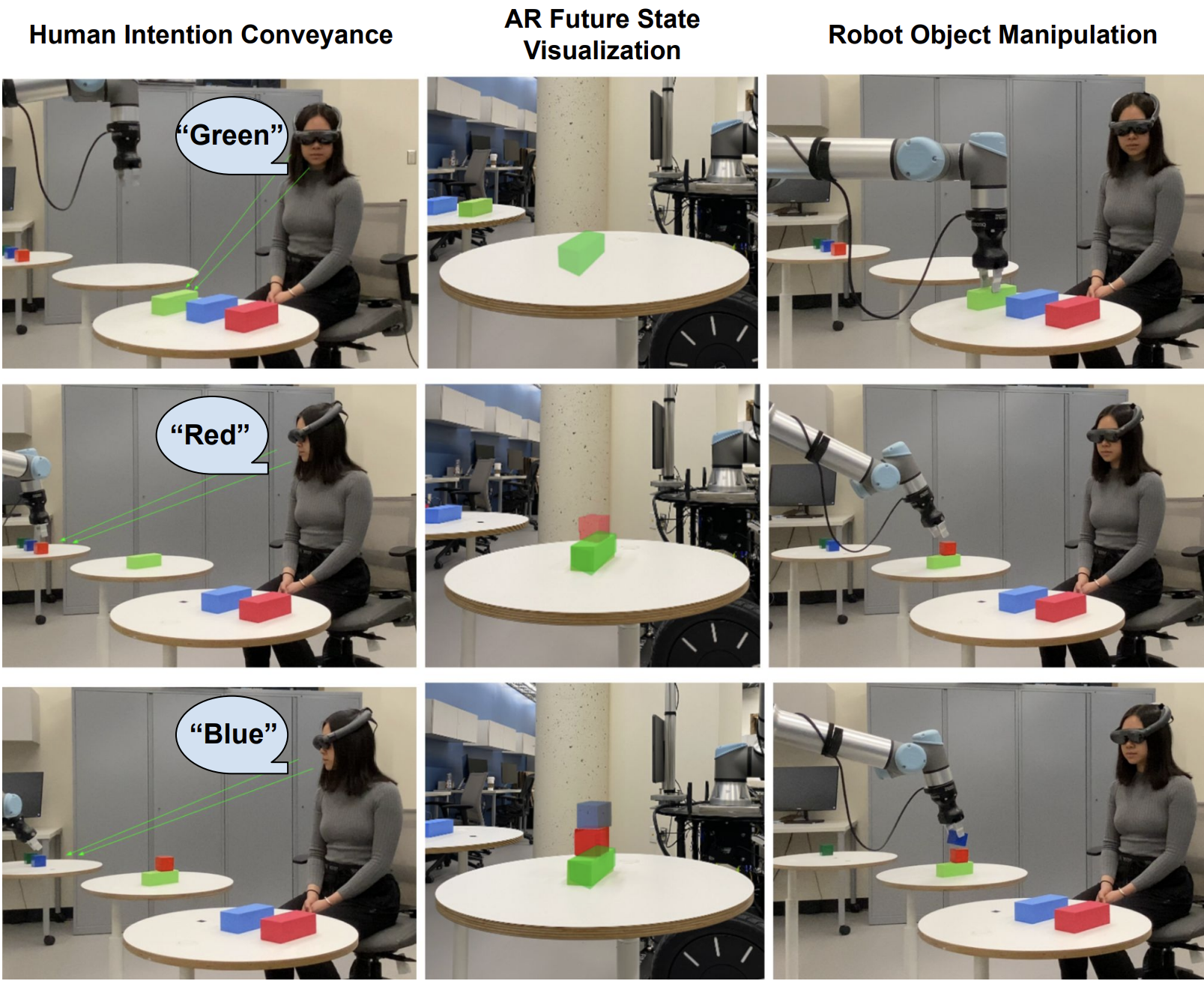}
  \caption{In an object stacking domain, the human specifies intent for the desired objects through natural language and eye gaze. ARDIE integrates these feedback modalities to display future states of the task in AR. When the human confirms the visualizations, a UR5e robot arm proceeds to object manipulation.}
  \vspace{-2em}
  \label{Fig1}
\end{figure}

Despite its advantages, current AR for HRC has several limitations. One challenge is the communication gap between humans and robots. Individual means of communication such as verbal cues may not be sufficient to effectively convey goals. Therefore, additional channels that can also reveal intentions, such as eye gaze, should be considered to improve bidirectional communication. Moreover, most AR research in HRC focuses on visualizing the present environment and robot status~\cite{muhammad2019creating}. However, it is equally crucial for humans to visualize future states of the environment in order to make current decisions. Our proposed system, ARDIE (AR with Dialogue and Eye Gaze), addresses this by providing real-time visual feedback of prospective outcomes to allow humans to make necessary adjustments before proceeding onto the next step.

We use a decision theoretic framework to model a joint policy that incorporates eye gaze data from a mixed reality (MR) headset, as well as language to enable intuitive communication. While gaze is oftentimes noisy and can only reference location or position, dialogue, on the other hand, can ground visual cues through responsive speech such as question asking and answering, which can further disambiguate between item references regardless of physical space. These two modalities used together can provide more effective communication. 
\begin{figure}[t]
  \includegraphics[width=\columnwidth]{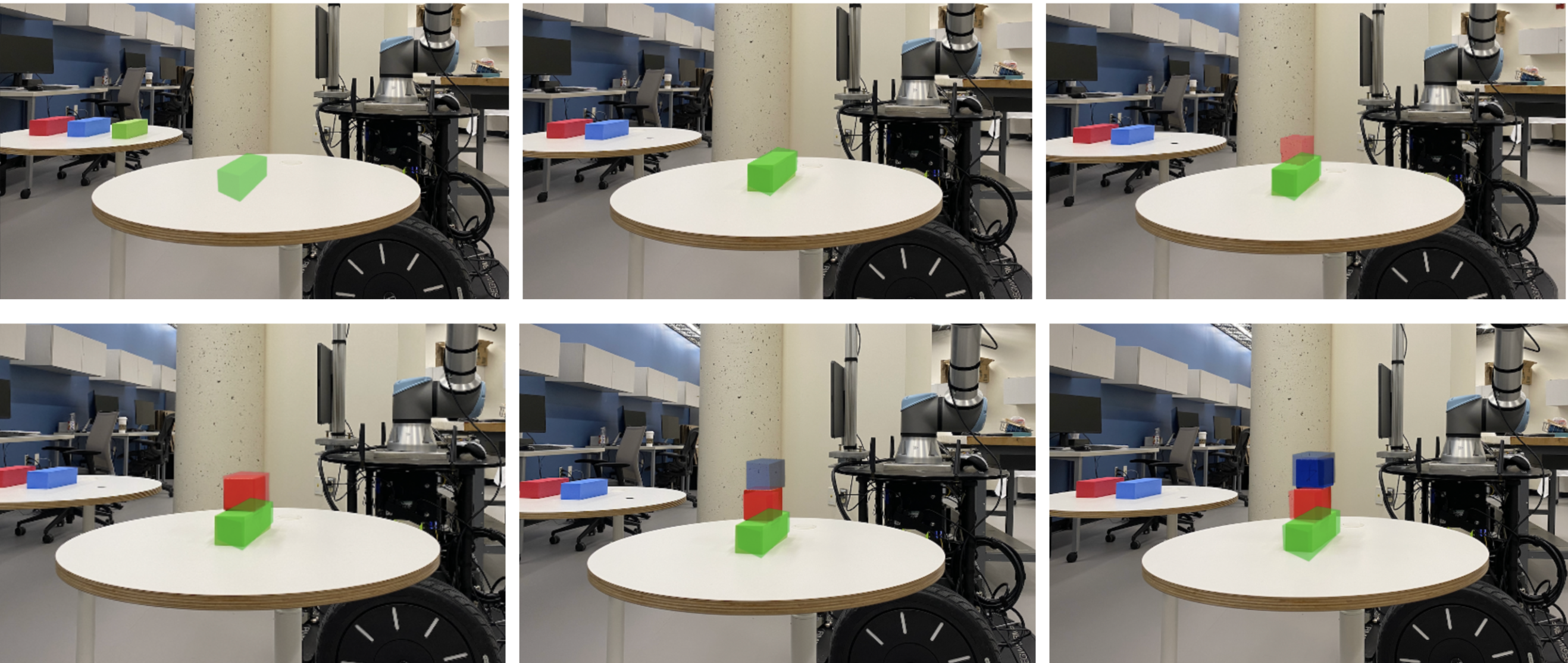}
  \caption{In this demonstration, the human wants to stack a large green block, a small red block, and a small blue block. For each block, the human conveys their intentions and an AR visualization is produced for the next state of the object sequence.}
  \vspace{-1em}
  \label{Fig 2}
\end{figure}

We formulate our problem under the scope of decision-making under uncertainty, where the state of the environment is not directly observable but can be inferred from a set of observations, and the actions of the agents can influence state transition and observations. Our joint policy specifies how the human and robot should act in each state of the environment based on the multi-modal feedback cues. While gaze is cheap and can reference various locations in space effectively, dialogue can further disambiguate implicit cues through direct speech and question answering. In turn, the joint output of our decision model provides optimal actions that can improve overall situational awareness.


\begin{figure*}[t]
\begin{center}

  \includegraphics[width=0.95\textwidth]{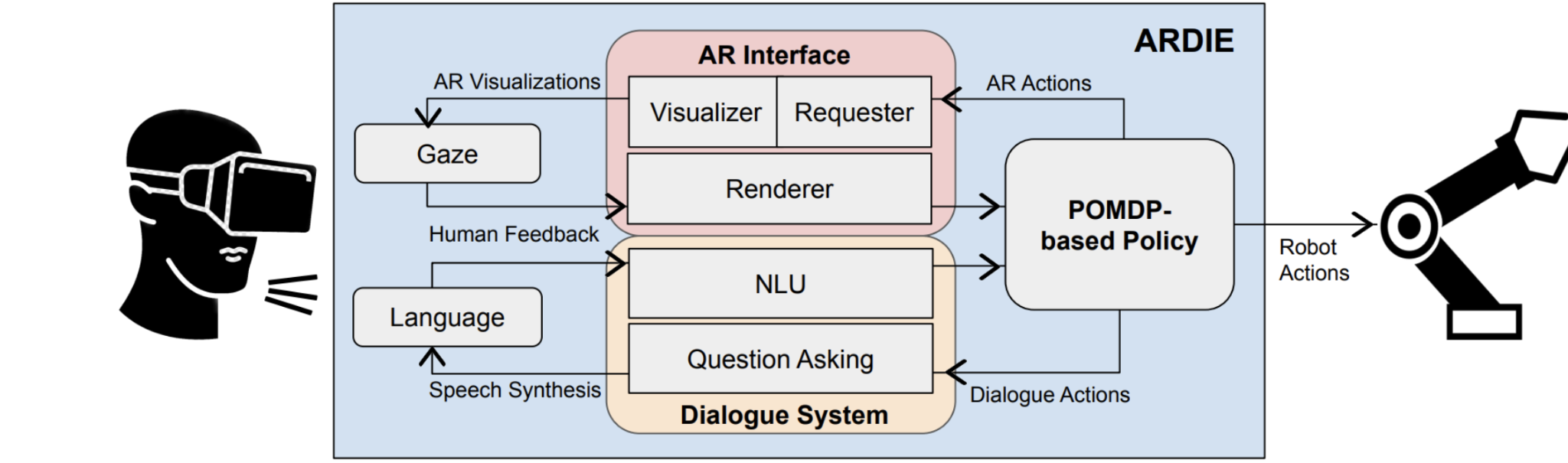}
  \caption{The human conveys intentions through gaze and language. The AR interface renders gaze as observation input into the POMDP planner. The dialogue system processes human speech through a natural language understanding (NLU) unit, and similarly, feeds that observation into the POMDP planner. The outputs of the planner are either a gaze request action, a future state AR visualization, a question asking action, or a robot action.}
  \vspace{-1.5em}
  \label{Fig 3}
  \end{center}
\end{figure*}
In this paper, we present ARDIE, a novel agent that enhances HRC by leveraging intuitive human feedback cues through a joint AR, dialogue, and eye gaze policy. Our system incorporates visualizations of current and future states to effectively convey intentions and improve decision-making by enhancing situational awareness. By providing a visual understanding of task outcomes, ARDIE enables users to adjust plans accordingly, leading to more effective collaboration. Our work provides several contributions. The first is a novel multi-modal agent system that takes in natural human feedback to convey future state visualizations for HRC tasks. Our second contribution is the implementation of this framework through an MR headset and a UR5e robot arm for an interactive object manipulation task.

\section{Related Works}
In this section, we discuss a few research areas that are relevant to our work on ARDIE.

\vspace{0.5em}
\noindent\textbf{AR for Robotics:}
Several studies have leveraged the capabilities of AR for robotics. In the past, researchers proposed a system that enables a mobile robot to project AR visual cues onto the floor, indicating intended actions to nearby humans \cite{chadalavada2015s}. There have also been proposed methods for visualizing intended assistance in shared control of robotic systems through trajectory overlays \cite{brooks2020visualization}, as well as telerobotic systems that allow human operators to remotely control a robot and receive visual feedback through head-mounted displays \cite{milgram1993applications}. Instead, we use an approach that leverages natural human feedback cues to provide a more intuitive HRC experience while our joint policy allows a more robust strategy to minimize ambiguity through the grounding provided by multiple feedback channels.

\vspace{0.5em}
\noindent\textbf{Natural Human Feedback Cues for HRC:}
Researchers have attempted to overcome the communication barrier between humans and robots by using natural language, where incorporating dialogue grounding into HRC systems can promote mutual understanding and clarification \cite{8542553}, \cite{amiri2019augmenting}. Overt eye gaze is shown to reveal human intent in directed attention tasks~\cite{eckstein2017beyond}, and gaze tracking has been previously integrated into robotic control systems \cite{7759741}, \cite{haji2018exploiting}. Despite these successes, single feedback channels may not be sufficient to resolve ambiguities occurring in noisy environments. For example, speech recognition is prone to translation errors and gaze may not be helpful in cluttered environments. ARDIE integrates both gaze and dialogue modalities with AR to adapt to dynamic environments in HRC.

\vspace{0.5em}
\noindent\textbf{Mixed Reality for Human Robot Communication:}
Researchers found that in an item disambiguation task, an MR interface to highlight tabletop items was most engaging, but the use of an AR projector was the most accurate \cite{8525554}. In another approach, the authors utilize an MR approach for human-robot communication by combining eye gaze, language, and pointing gestures to highlight and disambiguate between tabletop objects~\cite{inproceedings}. Instead, we leverage the benefits of both AR and human feedback cues for a human-robot \emph{collaboration} task, in which future states are shown using AR objects that possess physically plausible properties of the real object, such as structure, solid body, and shape. This component enables tasks that can not only distinguish between objects, but also tasks that require simulating object interactions, such as object stacking.

\vspace{0.5em}
\noindent\textbf{Internal State Visualizations:}
Researchers have developed interfaces that enable human workers to directly visualize internal states intentions of robots through AR \cite{muhammad2019creating}, \cite{chandan2021arroch}. However, in domains where human knowledge is essential in guiding robot tasks, it is critical for the human to be able to visualize not only the robot's intentions but also their own intended outcomes regarding the next decision step. ARDIE enables visualizations of human internal states by projecting the future state based on their current decisions. This allows humans to understand effects of their own actions to modify behaviors accordingly.

\section{Background on POMDPs}
A partially observable Markov decision process (POMDP) is a mathematical framework to model decision making in uncertain environments. Specifically, it is used in tasks where the agent only has partial information about the environment, and therefore must reason about its beliefs over possible states of the environment.
The system is modeled as a set of states $S$, observations $\Omega$, actions $A$, and rewards $R$, with the goal to learn an optimal policy $\pi$ that maximizes the expected cumulative reward over time. 

A POMDP is formalized as a tuple of $(S, A, \Omega, T, O, R, \gamma)$. At each time step, the environment is in some ground state $s \in S$, the agent takes an uncertain action $a \in A$ and receives an assigned reward $R(s,a)$. The system transitions to a new state $s'$, as modeled by the conditional probability transition function $T(s,a,s')$ = $p(s'|s,a)$. Next, an uncertain observation $z \in \Omega$ is received according to the observation function $O(s',a,z)$ = $p(z|s',a)$. Finally a discount factor $\gamma \in [0,1)$ is applied to ensure the reward is finite. Solving a POMDP produces a policy $\pi: B \mapsto A$ that maps a belief $b \in B$ to an action $a \in A$ that will maximize long-term rewards. A value function $V_{\pi}(b)$ that specifies the expected total reward of executing $\pi$ starting from $b$, can be approximately by a convex, piecewise-linear function:
\bigbreak
\begin{center}
\vspace{-1em}
$V(b) = \max\limits_{\alpha \in \Gamma} (\alpha \cdot b)$ 
\end{center}
\bigbreak
\noindent where $\Gamma$ is a finite set of vectors called $\alpha$-vectors, each associated with an action, $b$ is the discrete vector representation of a belief, and $\alpha \cdot b$ is the inner product of vectors $\alpha$-vector and $b$. The policy can be denoted by a set of $\alpha$-vectors. The agent begins with an initial belief $b_{0}$, and maintains a belief $b(s)$ about the state which is a probability distribution over all the possible states. Using Bayes rule, the belief $b(s)$ is updated when receiving observation $z$ after taking action $a$:
\bigbreak
\begin{center}
\vspace{-1em}
    $b'(s')$ = $\gamma O(z|s',a)\displaystyle \sum_{s \in S}^{}T(s'|s,a)b(s)$
\end{center}
\bigbreak 

\section{System Implementation}
Consider a simple object assembly task, where the human can select various objects with different properties from different locations, and a robot can help sequentially stack these objects in a target region. We model our task as a POMDP, where inputs are the user's intentions, inferred from eye gaze and dialogue responses. The system then generates AR visualizations of future states that align with human intent.

\subsection{AR Interface}
\vspace{-0.3em}
There are three major components to the AR interface. The \textbf{renderer} allows the human to display intent through eye gaze data for the different scene elements. When a gaze action is realized, the \textbf{requester} queries the human gaze through an MR headset, Magic Leap 1, to gather information regarding the approximate location of the physical object. Through the \textbf{visualizer}, when a visual action is realized, AR object representations are projected into the scene and modeled according to the real-world object by considering shape, solid body, mass, and gravity. 

\subsection{Dialogue System}
\vspace{-0.3em}
The dialogue system has two elements to allow the agent to strategically ask human questions and obtain spoken feedback. The \textbf{natural language understanding} (NLU) unit interprets human speech and utilizes a text-to-speech ROS package. The \textbf{question-asking} unit uses Google Cloud's speech-to-text API to allow the agent to query the human about the intended object or ask a confirmatory question.

\subsection{POMDP Planner}
\vspace{-0.3em}
We use the APPL offline POMDP planning software, based on the SARSOP algorithm for solving discrete POMDPs \cite{kurniawati2008sarsop}. Recall that a POMDP is modeled as a tuple $(S, A, \Omega, T, O, R, \gamma)$. Our system specifies the following:
\begin{itemize}
    \item  $S$: The state space is the set of objects that correspond to the humans intentions, where $S = \{s_1, s_2, ..., s_n\}$ for $n$ different intended objects in the environment.
    \item $A$: The set of actions correspond to different decisions the agent can take. Let $A$ = \{$a_{gaze}, a_{ask}, a_{project}$\} where $a_{gaze}$ requests the human to look towards the intended object, $ a_{ask}$ initiates the SDS to query the human about the object, and $a_{project}$ displays a visualization of the future state of the augmented environment, directly into the human's field of view. 
    \item  $\Omega$: The set of observations contains the partially observed states of the human's true intentions. Each observation $z \in \Omega$ corresponds to either a positional or descriptive property of the desired object, indicated by gaze and dialogue respectively.
    \item  $T$: The transition function $T(s,a,s')$ is an identity function for actions $a_{gaze}$ and $a_{ask}$ where the state remains unchanged. The $a_{project}$ actions are deterministic reporting actions that lead state transitions to terminate.
    \item  $O$: The observation function $O(s',a,z)$ models the noise in overt human gaze and spoken language. For $a_{gaze}$ and $a_{ask}$ we assume the probability of getting a correct observation is 0.8. 
    \item $R$: The reward function assigns the agent a large positive value for displaying the future state of the correct item, and penalizes the agent with a large negative value for an incorrect visualization. The gaze action is assigned a smaller negative value than the ask action.
    \item $\gamma$: The discount factor is set to 0.99 and determines the relative importance of future rewards compared to immediate rewards.
\end{itemize}

\subsection{Demonstration}
\vspace{-0.3em}
Consider a case where the human wants to visualize and stack three specific blocks sequentially, with a large green block on the bottom, then a small red block, and finally a small blue block on top. ARDIE begins by prompting the human to look at the first desired block to collect eye gaze position. Afterwards, ARDIE engages in a dialogue with the human to ask for the color. The human responds with "green", and after sufficient confidence, an AR representation of the desired object is projected onto a target location. ARDIE then asks a confirmatory question. If the configuration of the green block is confirmed, a robot action is sent to a UR5e arm to pick up and place the block accordingly. If the visualization does not look as intended to the human, the agent is re-instantiated to the last state, allowing the human to select another set of objects more appropriate for their goals. A similar process is done for the next small red block, and later the small blue block.

\section{Conclusion and Future Works}
We aim to bridge the communication gap in HRC tasks through ARDIE, a novel intelligent agent that utilizes a joint AR, dialogue, and eye gaze policy to improve situational awareness. ARDIE enables humans and robots to intuitively convey their internal state intentions and visualize current and future states of the environment. Our multi-modal system provides a more natural mechanism for humans and robots to communicate and make better informed decisions.

We realize a practical limitation of our system is the feasibility of scaling up to more complex and dynamic environments. One avenue for future work is to examine the physics of object interactions in greater depth to help model outcomes with better precision and accuracy. Incorporating more advanced physics engines can help humans and robots better understand the forces and constraints involved in object manipulation, leading to physically plausible behaviors in response to changes in the environment. Overall, we aim to progress efforts in improving coordination, communication, and collaboration between humans and robots.

\bibliographystyle{IEEEtran}
\bibliography{ref}

\begin{thebibliography}{10}
\providecommand{\url}[1]{#1}
\csname url@rmstyle\endcsname
\providecommand{\newblock}{\relax}
\providecommand{\bibinfo}[2]{#2}
\providecommand\BIBentrySTDinterwordspacing{\spaceskip=0pt\relax}
\providecommand\BIBentryALTinterwordstretchfactor{4}
\providecommand\BIBentryALTinterwordspacing{\spaceskip=\fontdimen2\font plus
\BIBentryALTinterwordstretchfactor\fontdimen3\font minus
  \fontdimen4\font\relax}
\providecommand\BIBforeignlanguage[2]{{%
\expandafter\ifx\csname l@#1\endcsname\relax
\typeout{** WARNING: IEEEtran.bst: No hyphenation pattern has been}%
\typeout{** loaded for the language `#1'. Using the pattern for}%
\typeout{** the default language instead.}%
\else
\language=\csname l@#1\endcsname
\fi
#2}}

\bibitem{dos2020situational}
C.~W. Dos~Santos, L.~Nelson~Filho, D.~B. Esp{\'\i}ndola, and S.~S. Botelho,
  ``Situational awareness oriented interfaces on human-robot interaction for
  industrial welding processes,'' \emph{IFAC-PapersOnLine}, vol.~53, no.~2, pp.
  10\,168--10\,173, 2020.

\bibitem{chandan2021arroch}
K.~Chandan, V.~Kudalkar, X.~Li, and S.~Zhang, ``Arroch: Augmented reality for
  robots collaborating with a human,'' in \emph{2021 IEEE International
  Conference on Robotics and Automation (ICRA)}.\hskip 1em plus 0.5em minus
  0.4em\relax IEEE, 2021, pp. 3787--3793.

\bibitem{bonarini2020communication}
A.~Bonarini, ``Communication in human-robot interaction,'' \emph{Current
  Robotics Reports}, vol.~1, pp. 279--285, 2020.

\bibitem{green2007augmented}
S.~A. Green, J.~G. Chase, M.~Billinghurst, and X.~Chen, \emph{Augmented reality
  for human-robot collaboration}.\hskip 1em plus 0.5em minus 0.4em\relax INTECH
  Open Access Publisher, 2007.

\bibitem{chandanlearning}
K.~D. Chandan, J.~Albertson, and S.~Zhang, ``Learning visualization policies of
  augmented reality for human-robot collaboration,'' in \emph{6th Annual
  Conference on Robot Learning}.

\bibitem{chen2019overview}
Y.~Chen, Q.~Wang, H.~Chen, X.~Song, H.~Tang, and M.~Tian, ``An overview of
  augmented reality technology,'' in \emph{Journal of Physics: Conference
  Series}, vol. 1237, no.~2.\hskip 1em plus 0.5em minus 0.4em\relax IOP
  Publishing, 2019, p. 022082.

\bibitem{azuma2001recent}
R.~Azuma, Y.~Baillot, R.~Behringer, S.~Feiner, S.~Julier, and B.~MacIntyre,
  ``Recent advances in augmented reality,'' \emph{IEEE computer graphics and
  applications}, vol.~21, no.~6, pp. 34--47, 2001.

\bibitem{muhammad2019creating}
F.~Muhammad, A.~Hassan, A.~Cleaver, and J.~Sinapov, ``Creating a shared reality
  with robots,'' in \emph{2019 14th ACM/IEEE International Conference on
  Human-Robot Interaction (HRI)}.\hskip 1em plus 0.5em minus 0.4em\relax IEEE,
  2019, pp. 614--615.

\bibitem{weisz2017assistive}
J.~Weisz, P.~K. Allen, A.~G. Barszap, and S.~S. Joshi, ``Assistive grasping
  with an augmented reality user interface,'' \emph{The International Journal
  of Robotics Research}, vol.~36, no. 5-7, pp. 543--562, 2017.

\bibitem{zhang2019vision}
Z.~Zhang, Y.~Li, J.~Guo, D.~Weng, Y.~Liu, and Y.~Wang, ``Vision-tangible
  interactive display method for mixed and virtual reality: Toward the
  human-centered editable reality,'' \emph{Journal of the Society for
  Information Display}, vol.~27, no.~2, pp. 72--84, 2019.

\bibitem{ghiringhelli2014interactive}
F.~Ghiringhelli, J.~Guzzi, G.~A. Di~Caro, V.~Caglioti, L.~M. Gambardella, and
  A.~Giusti, ``Interactive augmented reality for understanding and analyzing
  multi-robot systems,'' in \emph{2014 IEEE/RSJ International Conference on
  Intelligent Robots and Systems}.\hskip 1em plus 0.5em minus 0.4em\relax IEEE,
  2014, pp. 1195--1201.

\bibitem{chadalavada2015s}
R.~T. Chadalavada, H.~Andreasson, R.~Krug, and A.~J. Lilienthal, ``That's on my
  mind! robot to human intention communication through on-board projection on
  shared floor space,'' in \emph{2015 European Conference on Mobile Robots
  (ECMR)}.\hskip 1em plus 0.5em minus 0.4em\relax IEEE, 2015, pp. 1--6.

\bibitem{brooks2020visualization}
C.~Brooks and D.~Szafir, ``Visualization of intended assistance for acceptance
  of shared control,'' in \emph{2020 IEEE/RSJ International Conference on
  Intelligent Robots and Systems (IROS)}.\hskip 1em plus 0.5em minus
  0.4em\relax IEEE, 2020, pp. 11\,425--11\,430.

\bibitem{milgram1993applications}
P.~Milgram, S.~Zhai, D.~Drascic, and J.~Grodski, ``Applications of augmented
  reality for human-robot communication,'' in \emph{Proceedings of 1993
  IEEE/RSJ International Conference on Intelligent Robots and Systems
  (IROS'93)}, vol.~3.\hskip 1em plus 0.5em minus 0.4em\relax IEEE, 1993, pp.
  1467--1472.

\bibitem{8542553}
J.~Y. Chai, L.~She, R.~Fang, S.~Ottarson, C.~Littley, C.~Liu, and K.~Hanson,
  ``Collaborative effort towards common ground in situated human-robot
  dialogue,'' in \emph{2014 9th ACM/IEEE International Conference on
  Human-Robot Interaction (HRI)}, 2014, pp. 33--40.

\bibitem{amiri2019augmenting}
S.~Amiri, S.~Bajracharya, C.~Goktolgal, J.~Thomason, and S.~Zhang, ``Augmenting
  knowledge through statistical, goal-oriented human-robot dialog,'' in
  \emph{2019 IEEE/RSJ International Conference on Intelligent Robots and
  Systems (IROS)}.\hskip 1em plus 0.5em minus 0.4em\relax IEEE, 2019, pp.
  744--750.

\bibitem{eckstein2017beyond}
M.~K. Eckstein, B.~Guerra-Carrillo, A.~T.~M. Singley, and S.~A. Bunge, ``Beyond
  eye gaze: What else can eyetracking reveal about cognition and cognitive
  development?'' \emph{Developmental cognitive neuroscience}, vol.~25, pp.
  69--91, 2017.

\bibitem{7759741}
O.~Palinko, F.~Rea, G.~Sandini, and A.~Sciutti, ``Robot reading human gaze: Why
  eye tracking is better than head tracking for human-robot collaboration,'' in
  \emph{2016 IEEE/RSJ International Conference on Intelligent Robots and
  Systems (IROS)}, 2016, pp. 5048--5054.

\bibitem{haji2018exploiting}
A.~Haji~Fathaliyan, X.~Wang, and V.~J. Santos, ``Exploiting three-dimensional
  gaze tracking for action recognition during bimanual manipulation to enhance
  human--robot collaboration,'' \emph{Frontiers in Robotics and AI}, vol.~5,
  p.~25, 2018.

\bibitem{8525554}
E.~Sibirtseva, D.~Kontogiorgos, O.~Nykvist, H.~Karaoguz, I.~Leite,
  J.~Gustafson, and D.~Kragic, ``A comparison of visualisation methods for
  disambiguating verbal requests in human-robot interaction,'' in \emph{2018
  27th IEEE International Symposium on Robot and Human Interactive
  Communication (RO-MAN)}, 2018, pp. 43--50.

\bibitem{inproceedings}
E.~Rosen, D.~Whitney, M.~Fishman, D.~Ullman, and S.~Tellex, ``Mixed reality as
  a bidirectional communication interface for human-robot interaction,'' 09
  2020.

\bibitem{kurniawati2008sarsop}
H.~Kurniawati, D.~Hsu, and W.~S. Lee, ``Sarsop: Efficient point-based pomdp
  planning by approximating optimally reachable belief spaces.'' in
  \emph{Robotics: Science and systems}, vol. 2008.\hskip 1em plus 0.5em minus
  0.4em\relax Citeseer, 2008.

\end{thebibliography}

\end{document}